# Practical Issues in Constructing a Bayes' Belief Network

Max Henrion,
Department of Engineering and Public Policy,
and Department of Social and Decision Science,
Carnegie Mellon University, Pittsburgh, Pa 15213.

## Abstract

Bayes belief networks and influence diagrams are tools for constructing coherent probabilistic representations of uncertain expert opinion. The construction of such a network with about 30 nodes is used to illustrate a variety of techniques which can facilitate the process of structuring and quantifying uncertain relationships. These include some generalizations of the "noisy OR gate" concept. Sensitivity analysis of generic elements of Bayes' networks provides insight into when rough probability assessments are sufficient and when greater precision may be important.

## 1. Introduction

As the advantages of coherent probabilistic schemes for representing uncertainty become more widely recognized, there is increasing interest in the use of Bayesian belief networks and influence diagrams (Howard & Matheson, 1984, Pearl, 1986, Shachter, 1987, Henrion, 1987). Influence diagrams have been in use for several years by decision analysts as a tool to help in constructing simple models (typically with less than 8 uncertain and decision variables), that can be analyzed by decision trees. However, to date there has been a lack of published examples of applications employing general networks which are significantly larger. Doubtless, this is due in part to the lack of availability until recently of practical algorithms for propagating evidence through complex networks. It may also be due in part to the difficulties of the knowledge engineering: As always, there is a lag between the development of the theory and the development of practical techniques and skills for applying them.

The goal of this paper is to discuss some techniques which can ease the process of structuring large networks and quantifying probabilistic influences. These will be illustrated by their use in the construction of an actual belief network and influence diagram of moderate size (30 variables). It will use this example to illustrate a variety of general techniques for conducting and facilitating the knowledge engineering process. It will also examine some general issues of analyzing the sensitivity of conclusions to errors and approximations in assessed probabilities. This helps to provide insights about which kinds of errors in numerical probability assessments are or are not likely to make much difference to the results.

## 2. The task

The task involved construction of a system to aid in the diagnosis and selection of treatment for root disorders of apple trees. This involved the encoding of the expert knowledge of a plant pathologist with ten years experience as a consultant to orchardists on these problems. This work was part of an experimental study to compare the construction of a rule-based expert system with a decision analytic/belief net model for the same task. The comparison of these two approaches is reported elsewhere (Henrion & Cooley, 1987). This paper will address only the latter approach.

There are several possible causes of root damage to apple trees, including water stress from waterlogged soil, cold stress from severe fall or winter, and infection by the fungus, *phytophthora*. These problems often lead to damage and destruction of apple trees, and so are of major commercial significance to

132

orchardists. Abiotic stresses from excess water or cold can increase susceptibility to phytophthora infection, as well as causing root damage themselves. Moderate cases of phytophthora can be controlled by applying a fungicide. Other treatments include tiling and draining the area to control water damage, and bridge-grafting. The plant pathologist uses a wide variety of evidence to diagnose the cause of root damage, and so recommend treatments. These include information about the tree, e.g. whether its root stock is resistant to phytopthora, environmental conditions, e.g. recent rain, observable symptoms, e.g. root cankers, and laboratory tests. (See Figure 1.)

## 3. Structuring the network

The initial focus was on the decision whether to treat the tree with a fungicide which may control a phytophthora infection. Choices during the knowledge engineering about which elements to include in the model and which to ignore were made according to whether they seemed likely to affect this decision. The first part of the model to be elicited was the outcome value, expressed as the total dollar cost to the orchardist, as affected by the decision and the factors that might affect this, including the value of the tree, cost of treatment, and tree damage due to the various disorders. These appear in the lower part of Figure 1.

The diagnostic part of the network links the observable data (indicants) with the hypothesized, but unobservable root disorders. This was elicited progressively from the expert, starting with the disorders, using questions like "What could affect this? What observable results could this have? What evidence would you look for?" Each link was assigned an arrow according the direction of causal influence as judged by the expert. The initial structuring segment of the interview took about a day. The final results are shown in Figure 1.

## 4. Defining levels for variables

Once an initial network was constructed, the next stage was to define the number of levels for each variable. Most of the variables are intrinsically continuous, but all were discretized, most as binary variables, but a few with three or four levels. In general, those variables which seemed most likely to affect the outcome and hence the decision were modelled in more detail. For example, Eventual-tree-damage was assigned four levels {None, Temporary-damage, Permanent-damage, Tree-needs-replacing}, and the main disorders, including current-Phytophthora-damage and Abiotic-Stress were each assigned three levels, {None, Recoverable, Beyond-recovery}. The intuitions of the expert about what he finds important were also used to help make these decisions about discretization. Since the complexity of the conditional probability distributions needed to express the uncertain influences increases rapidly with the number of levels, it is critical to keep them well under control.

Each level of each quantity was carefully defined. For example, Temporary-damage, was defined as 25% reduction in fruit production this year, and Permanent-damage was defined as 60% reduction this year, and 15% in all future years. Any tree is deemed uneconomic, to be uprooted and replaced, if its productivity is permanently impaired by more than 25%. Such explicit definitions are necessary to avoid ambiguity when assessing conditional probabilities.

## 5. Quantification of influences

Influences were quantified as conditional probability distributions. Initially qualitative judgments were sought about the strength of the influences and the relative impacts of different parents where there were multiple causative factors. Generally, each conditional probability was first expressed by a phrase, such as "a toss-up", "unlikely", or "impossible". Then each was quantified by an approximate number, generally from the set {0, .01, .05, .1, .2, .3, .5, .7, .8, .9, .95, .99, 1}

133

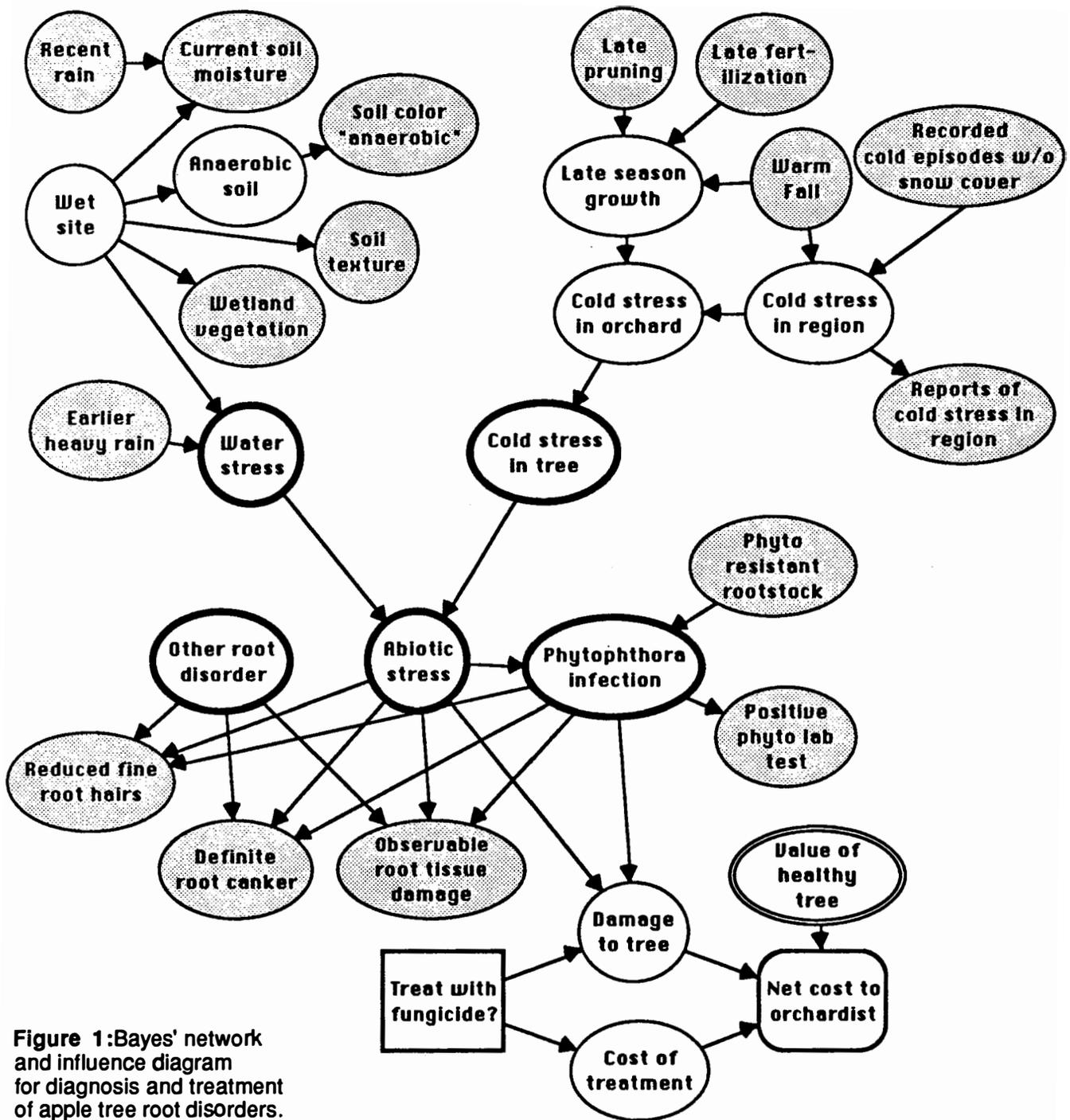

**Figure 1:** Bayes' network and influence diagram for diagnosis and treatment of apple tree root disorders.
Prepared using David (Shachter, 1986)

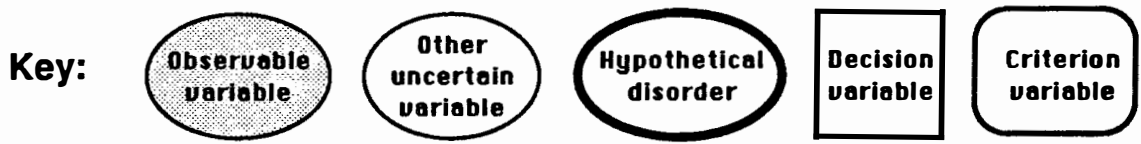

Key: Observable variable | Other uncertain variable | Hypothetical disorder | Decision variable | Criterion variable

## 6. Refinement of model structure

During the quantification of influences there was also further refinement of the network whenever it turned out that modifications to the structure could make it easier to assess, or earlier structural assumptions turned out to be mistaken. One important example concerned violations of conditional independence. By the definition of a Bayes network the successors of a node should be conditionally independent given that node. If, on further reflection, they turn out not to be, then an additional node may be required to fix the problem. For example, initially the expert judged that Cold Stress in Orchard and Reports of Cold Stress in Region were both causally influenced by Records of Winter Cold Episodes without Snow cover (Figure 2a). But the expert later judged that they weren't conditionally independent of it. However, further discussion of the issues revealed that there was a hidden hypothesis that there actually was Cold Stress in the Region. (See Figure 2b.) This could not be observed directly, but might lead to Cold Stress in the Orchard and to Reports of Cold Stress in the Region, and the latter two *were* judged conditionally independent on this variable. As Pearl has suggested, conditional independence should not be viewed as an awkward constraint on the freedom of the knowledge engineer, but rather as a useful aid. Where it fails, this is an indication that there is a hidden variable, whose introduction may greatly ease the assessment process.

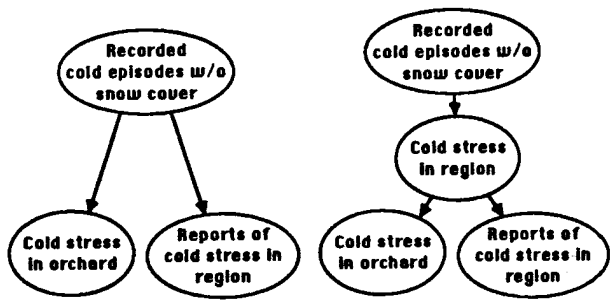

**Figure 2a:** Initial network violating conditional independence.

**Figure 2b:** Additional node restores conditional independence

Another way in which an additional node may simplify the assessment process is where two factors were judged to have identical effects if either or both were present. For example, Water-Stress and Winter-Stress at a given level were both judged to have the same impact on each of the three symptoms, Reduced-Root-Hairs, Canker-Margin, Root-Tissue-Damage, and on increasing susceptibility to Phytophthora. (See Figure 1). Thus an additional variable was defined, Abiotic-Stress, which is the conjunction of Water-Stress and Winter-Stress. (Actually, the level of Abiotic-Stress is defined as the maximum of the levels of these two.) Since the size of each conditional distribution is multiplied by the number of levels of each causal factor (three in this case), this replacement produced a useful factor of three reduction in the difficulty of assessing the influences on each of the four impacted variables.

## 7. Noisy OR gates

Considerable reduction in effort was also obtained by using versions of the "noisy OR gate" (Pearl, 1986) generalized for n-ary variables. The "noisy OR" influence structure applies when there are several possible logical causes, $x_i$, $i=1,2,...n$ of a binary effect variable $y$, where (a) each of which has a probability $p_i$ of being sufficient to produce the effect in the absence of all other causes, and (b) the probability of each cause being sufficient is independent of the presence of other causes. Pearl shows that the probability of $y$ given a subset $X$ of the $x_i$ which are present:

$$p(y|X) = 1 - \prod_{i:\, x_i \in X} (1-p_i)$$

The complete conditional distribution for $n$ binary predecessors would require the specification of $2^n$ parameters, but if the noisy OR assumptions apply, you need specify only $n$, one for each $p_i$.

Like any model, a Bayes' net is never complete, and so there will often be possible causes of an effect that are not explicitly modelled. To allow for this in a noisy OR it is useful to assess a *base rate probability*, $p_0$ for "all other causes", i.e. the event that the effect will occur apparently spontaneously in the absence of any of the causes modelled explicitly. If the expert assesses the probability that each explicit predecessor is sufficient to cause the effect variable in the absence of any other explicit cause, the numbers need readjustment to obtain the formula for the probability of $y$:

$$p(y|X) = 1 - (1-p_0) \cdot \prod_{i \cdot x_i \in X} \frac{1-p_i}{1-p_0}$$

A potential example of a noisy OR gate with a base rate probability occurs in Late-Season-Growth, of the apple tree, (which renders the tree more susceptible to Cold-Stress) which is influenced by Late-Pruning, Late-Fertilization, or a Warm-Fall. (See right top of Figure 1.) Originally this influence was assessed completely by obtaining the probability of the effect conditional on each of the eight combinations of its three binary causal influences. The distribution obtained was:

| Pruning | No | | Yes | |
|---|---|---|---|---|
| Warm fall | No | Yes | No | Yes |
| Late     No | .1 | .6 | .8 | .9 |
| Fertil  Yes | .8 | .9 | .9 | 1 |

Instead, this could have been modelled as a noisy OR with base rate $p_0=0.1$, and sufficiency probabilities for each of the three possible causes alone to match the three specified. This would have required only four instead of eight numerical assessments. The resulting conditional distribution is almost identical for most purposes:

| Pruning | No | | Yes | |
|---|---|---|---|---|
| Warm fall | No | Yes | No | Yes |
| Late     No | .1 | .6 | .8 | .91 |
| Fertil  Yes | .8 | .91 | .96 | .98 |

The noisy OR influence can be generalized to apply to cases with multi-valued variables. For example, the observable symptom reduced fine root hairs (binary) can be produced by any of the three disorders Other-root-problems, Abiotic-stress, or Phytopthora, the latter two of which have three different levels. The entire conditional distribution has 2x3x3 =18 independent parameters. If we assume that the probability that each level of each disorder is sufficient to cause Reduced-fine-root-hairs is independent of the other two disorders, then we have a simple extension of a noisy OR. In this case Other disorders is modelled explicitly, and so the base probability in the absence of any of the three classes of disorder is zero. The full distribution may be assessed given the probability that each of the levels other than "none" of each input is sufficient to cause the symptom, i.e. 1+2+2=5 parameters. Again the conditional distribution generated by these was very similar to the numbers actually assessed. The largest difference was 0.15, and the standard deviation was 0.06.

The notion of the noisy OR can be further generalized to deal with effect variables with multiple levels. For example, consider the symptom "observable tissue-damage", which has four levels, {none, low, moderate, severe}, and is influenced by the same three disorders as "reduced fine root hairs". In effect one can treat an $n$-ary variable as $n-1$ binary variables. The resulting value for the effect variable is the *maximum* of the levels produced by each of its influencing variables. Thus for example, Observable-tissue-damage is assumed to be severe if any of the three disorders is sufficient to cause severe damage. Notice, that no interaction is assumed, so that if all three disorders are sufficient to cause moderate damage, then the net result is moderate. In this case the complete conditional distribution has 3x3x3x2=54 free parameters, but with this extended noisy OR assumption, only 10 numbers needed to be assessed. Again the probabilities generated from these were very similar to those assessed directly.

## 8. Model testing and refinement

A set of 8 typical orchard scenarios, each with a set of observed values for some or all of the indicant variables were constructed by the expert for testing of the system. These

136

include some which are relatively easy with strong evidence for a single diagnosis, and some which are harder with weak or conflicting evidence. For the easy cases, we should certainly expect the system to come up with a similar conclusion to the expert. But for the hard cases, they may come to different conclusions, either because the expert is using knowledge not properly represented in the model, or because the expert's reasoning under uncertainty is faulty relative to the system, which is based on a normative theory. The findings of behavioral decision theory suggest that human judgment under uncertainty will typically be error-prone where we must reason beyond direct experience. This study revealed a clear example of such a conflict between the intuition of the expert and the inferences of the system: The expert's judgment about the expected value of the fungicide treatment for some typical scenarios turned out to be a significant overestimate. Detailed examination and explanation of the reasoning of the system convinced the expert that the system was in fact correct, and consequently he gained new insight into the problem and modified his intuition. This case is described in more detail in (Henrion & Cooley, 1987).

## 9. Sensitivity analysis

Sensitivity analysis is useful to discover the relative importance of different indicants (symptoms, environmental conditions and other observable factors) in arriving at a diagnosis and hence in selecting the decision. It is also useful during the construction of the network and quantification of influences to help identify what parts of the model and which parameters may be critical and which are less likely to be important to the results. In this way it can provide a guide for allocation of the effort in knowledge engineering towards the most important elements.

One useful measure of sensitivity to predictive or causal evidence is the *sensitivity range* of the probability of an event $y$ with respect to an event $x$. Suppose $e$ is an assessment error (viewed as an event) which might affect the assessment of the probability of $x$, giving $p(x|e)$. Suppose that $y$ is conditionally independent of $e$ given $x$. Then the sensitivity range is defined as the derivative of $p(y|e)$ with respect to $p(x|e)$:

$$SR(y,x) \equiv \frac{dp(y|e)}{dp(x|e)}$$

Given conditional independence, so that $p(y|x) = p(y|x,e)$, we have,

$$p(y|e) = p(y|x).p(x|e) + p(y|\overline{x}).(1-p(x|e)).$$

Taking the derivative with respect to $p(x|e)$, we get,

$$SR(y,x) = p(y|x) - p(y|\overline{x}). \quad [1]$$

So, it turns out that the sensitivity range is equal to the maximum possible change in the probability of $y$ which can be caused by error $e$ as $p(x|e)$ varies from 0 to 1.

Since a sensitivity range is a difference between two probabilities, its absolute magnitude cannot be greater than one. If the link is non-deterministic, then it must be strictly less than one:

$$|SR(y,x)| < 1 \quad [2]$$

Consequently the effect of any error in judging a probability of a causal or predictive variable cannot be greater than the magnitude of the error. Thus errors in probability assessment cannot be "amplified" by the model. In general they will be attenuated as the number of cascaded uncertain inferences is increased. Suppose we have a causal chain:

$$e \longrightarrow x_1 \longrightarrow x_2 - \longrightarrow x_n$$

It is easy to show that the sensitivity range of the end of a chain with respect to an error in the probability $p(x_1|e)$ is simply the product of the sensitivity ranges for each of the intermediate links:

$$SR(x_n, x_1) = \prod_{i}^{n-1} SR(x_{i+1}, x_i). \quad [3]$$

We may also consider the potential impact in errors in assessments of the link probabilities, $p(x_{i+1}|x_i)$. Since the absolute sensitivity range for each link must be less than unity if it is non-deterministic, each link serves to dilute the impact of all the others, whether they come before or after it in the chain. This supports the

137

intuition that the longer a chain of uncertain reasoning, the more tenuous the results. For example, consider the upper right-hand corner of the influence diagram in Figure 1. Late pruning could lead to late season growth, which could in turn lead to cold stress in the orchard, and hence to cold stress in the tree. Each step adds additional uncertainty, and the resulting degree of belief in cold stress in the tree will not be very sensitive to changes in the prior on late pruning or on any of the link probabilities.

However, things can be a little different for diagnostic links. Suppose $A$ influences $B$,

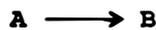

Suppose we define the odds of $a$ as $O(a)$, and the likelihood ratio:

$$L(b,a) = \frac{p(b|a)}{p(b|\bar{a})}.$$

We can then restate Bayes' theorem in this form:

$$p(a|b) = \frac{L(b,a).O(a)}{L(b,a).O(a) + 1}. \quad [6]$$

We take the derivative of this posterior with respect to the Likelihood ratio as a measure of sensitivity, getting:

$$\frac{dp(a|b)}{dL(b,a)} = \frac{O(a)}{(L(b,a).O(a) + 1)^2}. \quad [7]$$

This sensitivity can get large when $O(a)$ is large and $L(b,a)$ is small.

We can illustrate this from the apple tree example with the relationship between Cold stress in region and reports of cold stress in region, which we will identify as $A$ and $\bar{B}$ respectively. (Note that $\bar{B}$ means *no* reports.) Suppose we already know information $c$, that there was a warm Fall and that there were recorded cold episodes without snow cover, but have not yet sought any other information. According to the probabilities provided by the expert, cold stress in the region is almost certain, $p(a|c) = 0.95$. The expert also judged that the probability of no reports given cold stress in the region is extremely small, and the probability of reports given no cold stress is very large: $p(b|a) = 0.025, p(b|\bar{a}) = 0.95$. Hence the Likelihood ratio is small, $L(b,a) = 0.026$. If, when we ask, we find there have been no reports of cold stress, then from [7] the posterior for cold stress is less likely than not, $p(a|b) = 0.33$. On the other hand, if the expert had judged the probability of no reports given cold stress to be a little higher, say $p(b|a) = 0.1$, then the posterior would be more likely than not, $p(a|b) = 0.67$. In other words a 0.075 increase in the conditional probability (and similar increase in likelihood ratio) leads to a 0.33 increase in the posterior, a sensitivity factor of 4.4. It is interesting to note that this sensitivity arises where two sources of evidence are in conflict (in this case the weather history conflicts with the absence of reports of cold stress). If both sources supported the same conclusion it would not occur.

We should also recognize that this small increase in absolute terms, corresponds to a factor of 4 increase in relative terms. The odds-likelihood form of Bayes' theorem shows that this fourfold increase in the likelihood leads to a fourfold increase in the posterior odds. A good argument can be made that the magnitude of errors or vagueness in assessed probabilities is not uniform for probabilities between 0 and 1. Uniform vagueness in log-odds seems more reasonable. Applying this transform to Bayes' theorem we obtain the posterior log-odds as the sum of the prior log-odds and the log-likelihood ratio:

$$Ln[O(a|b)] = Ln[O(a)] + Ln[L(b,a)]$$

In this metric the sensitivity of the posterior to error in the link strength (expressed as log-likelihood or evidence weight), is always unity. Sensitivity to error in the prior is also unity. This discussion also underlines the importance of the distinction between almost certain and absolutely certain: Assigning 0 or 1 can have a drastically different effect than assigning 0.01 or 0.99 say.

## 10. Conclusions

This example at least demonstrates the feasibility of the knowledge engineering in constructing a Bayes' net or influence diagram of moderate size. Like more conventional knowledge engineering with rule-based inference schemes, it is a demanding process. However, I have used it to illustrate a number of techniques which can make it considerably easier both for the knowledge engineer and the domain expert, and which may help to improve the reliability of the result.

There is no doubt still considerable scope for developing additional skills and techniques to further facilitate the process. In particular, there are a number of ways in which improved software tools could help greatly. Graphic programs for entering and editing influence diagrams directly, such as David (Shachter, 1986) and Demaps (Wiecha, 1986) can considerably facilitate the structuring process. Provision of built-in facilities for specifying noisy OR gates and their extensions could further speed the encoding of influences with many parent variables. Still more useful would be facilities to support dynamic sensitivity analysis of results to uncertainty and vagueness in assessed probabilities for plausible ranges of scenarios, to guide the construction process to focus on those parts of the model that are demonstrably most critical.

### Acknowledgements

I am very grateful to Daniel Cooley, who contributed his expertise in plant pathology with patience and good humor. This work was supported by the National Science Foundation under grant IST-8603493 to Carnegie Mellon.

### References

Henrion, M. (1987). Uncertainty in Artificial Intelligence: Is probability epistemologically and heuristically adequate? In J. Mumpower (Ed.), *Expert Systems and Expert Judgment*. NATO ISI Series: Springer-Verlag.

Henrion, M. & Cooley, D.R. (1987). An Experimental Comparison of Knowledge Engineering for Expert Systems and for Decision Analysis. *Proceedings of AAAI-87: Sixth National Conference on Artificial Intelligence, Seattle, Wa.*, AAAI, Los Altos, Ca.

Howard, R.A. & J. E. Matheson. (1984). Influence Diagrams. In Howard, R.A. & J. E. Matheson (Eds.), *The Principles and Applications of Decision Analysis: Vol II*. Strategic Decisions Group, Menlo Park, CA.

Pearl, J. (1986). Fusion, Propagation, and Structuring in Belief Networks. *Artificial Intelligence, 29*(3), 241-288.

Shachter, R.D. . (1986). David: Influence Diagram Processing System for the Macintosh. In *Proceedings of Second Workshop on Uncertainty in Artificial Intelligence*. AAAI.

Shachter, R.D. (1987). Evaluating Influence Diagrams. *Operations Research*, Vol. April.

Wiecha, C. (1986). *An empirical study of how visual programming aids in comprehending quantitative policy models*. Doctoral dissertation, Carnegie Mellon University, Department of Engineering and Public Policy,